# Quantitative Results Comparing Three Intelligent Interfaces for Information Capture: A Case Study Adding Name Information into an Electronic Personal Organizer

**Jeffrey C. Schlimmer** SCHLIMME@EECS.WSU.EDU
*School of Electrical Engineering & Computer Science*
*Washington State University, Pullman, WA 99164-2752, U.S.A.*

**Patricia Crane Wells** PATRICIA@ALLPEN.COM
*AllPen Software, Inc.*
*16795 Lark Avenue, Suite 200, Los Gatos, CA 95030, U.S.A.*

## Abstract

Efficiently entering information into a computer is key to enjoying the benefits of computing. This paper describes three intelligent user interfaces: handwriting recognition, adaptive menus, and predictive fillin. In the context of adding a person's name and address to an electronic organizer, tests show handwriting recognition is slower than typing on an on-screen, soft keyboard, while adaptive menus and predictive fillin can be twice as fast. This paper also presents strategies for applying these three interfaces to other information collection domains.

## 1. Introduction

When you meet someone new, you often wish to get their name and phone number. You may write this in a small notebook or personal organizer. This takes a few minutes to do, so you put their business card or a small slip of paper in your organizer, promising to copy it over at a later time.[1] When that later time comes, you face the tedious task of finding where the now-several names should go in your organizer and recopying the information. If you are comfortable with computers, you may use an electronic organizer (a small computer that includes software for managing names and appointments). Looking up someone's phone number is faster with these devices, but adding is more tedious, and owning is more costly. As a concession to reality, these devices often include pockets for holding queued slips of paper.

What solutions could we propose to eliminate this procrastination? If adding a person's name[2] to your organizer were fun (say with your choice of inspirational message, gratuitous violence, or a lottery ticket), you might add names more readily. Avoiding this whimsy, we could get the desired effect by just making it faster to add a person's name. For this reason paper organizers use index tabs; electronic organizers use automatic filing. To be faster still, an organizer could read a card or handwritten note (via optical character or handwriting

1. We have a friend who places a rubber band around his organizer to ensure that these paper slips don't escape before their time.
2. For brevity, we'll refer to a person's name, address, phone numbers, e-mail, etc. as a "name". Whether the aggregate information or just a person's first or last name is intended should be clear from context.

recognition). Applying artificial intelligence ideas, we could even imagine an organizer that predicts what you need to write and does it for you.

This paper describes an electronic organizer that is almost as fast as a slip of paper, and certainly much faster than previous organizers. It uses commercial hardware (Newton, described in Section 2). Its software has three interface components designed to speed adding a person's name (described in Section 3): handwriting recognition, adaptive menus with recent values, and predictive fillin. The primary contributions of this paper are detailed evaluations of the benefits of these three components (described in Section 4).

Adding a person's name into an organizer is a special case of capturing and organizing information. This is a ubiquitous task. Big businesses institute careful procedures with custom forms and databases, but there are billions of smaller, one-or-two-person tasks which could be done more efficiently and accurately if getting information into a computer were easier. Even small gains would be repeated many times over whenever someone needed to collect information to make a decision, monitor a process, or investigate something new. The secondary contributions of this paper are a consideration of how the three components may be applied more broadly (described in Section 5).

These three interface components are robust if familiar. Whether they are "intelligent" is arguable. Some advocate a behavior-based definition which may also apply here, i.e., that the question of whether a device is intelligence or not should be answered by examining its behavior rather than its internal processes and representations (e.g., Agre & Chapman, 1987; Horswill & Brooks, 1988). Even if it does not, our goal is to address the question of how much intelligence, agency, or support one wants in an interface (Lee, 1990; Rissland, 1984). We assert: as much as will speed the user's performance of the task. Furthermore, much research is directed at automatically learning what we will hard-code in this study (e.g., Dent, Boticario, McDermott, Mitchell, & Zabowski, 1992; Hermens & Schlimmer, 1994; Schlimmer & Hermens, 1993; Yoshida, 1994). Even if learning works perfectly, is the result worthwhile? We claim the answer can be found in an empirical study of the usefulness of various user interface components.

## 2. Newton

Newton is an operating system introduced by Apple Computer, Inc. in 1993. It was designed for a single-user, highly-portable computer. Frames are the central data structure in Newton. They are stored in persistent object databases maintained in RAM (Smith, 1994).

Each Newton computer includes a pressure-sensitive, bitmapped display on which the user writes, draws, or taps to enter information (Culbert, 1994). All are small enough to hold in one hand and weigh around one US pound. Battery life is about one day's worth of continuous use. For a thorough overview of the hardware and software context of current pen computers, the reader may wish to consult (Meyer, 1995).

From the very lowest levels, Newton supports recognition. Its handwriting recognition was highly publicized when first introduced. The recognizer allows free-form input of printed and cursive writing.[3] It uses on-line recognition to convert writing to Unicode[4] text. The recognizer uses contextual information to limit both types of characters within specific fields and combinations of characters within words. The latter relies heavily on a

3. Throughout this paper references to handwriting also refer to handprinting. Where a distinction is required, the latter term will be used explicitly.

4. A character encoding similar to ASCII but with two-bytes per character for languages with larger character sets.

dictionary—only words appearing in the dictionary can be recognized. If the user types a new word using an on-screen soft keyboard,[5] Newton volunteers to add it to the dictionary for future recognition. Optionally, the user can invoke a secondary recognizer that does not use a dictionary and attempts to recognize what is written "letter by letter"; Section 4 describes the accuracy of this option. Application developers can customize handwriting recognition by providing special purpose dictionaries or a regular expression describing the syntax of words to be recognized.

Most Newton computers include several applications in ROM so they can serve as an electronic organizer. Relevant to the point of this paper, all Newton computers to date include an application called "Names" for storing and retrieving people's names, addresses, etc.; Figure 1 depicts this application. Section 4 describes experiments using a standard and enhanced version of the Names application to add people's names.

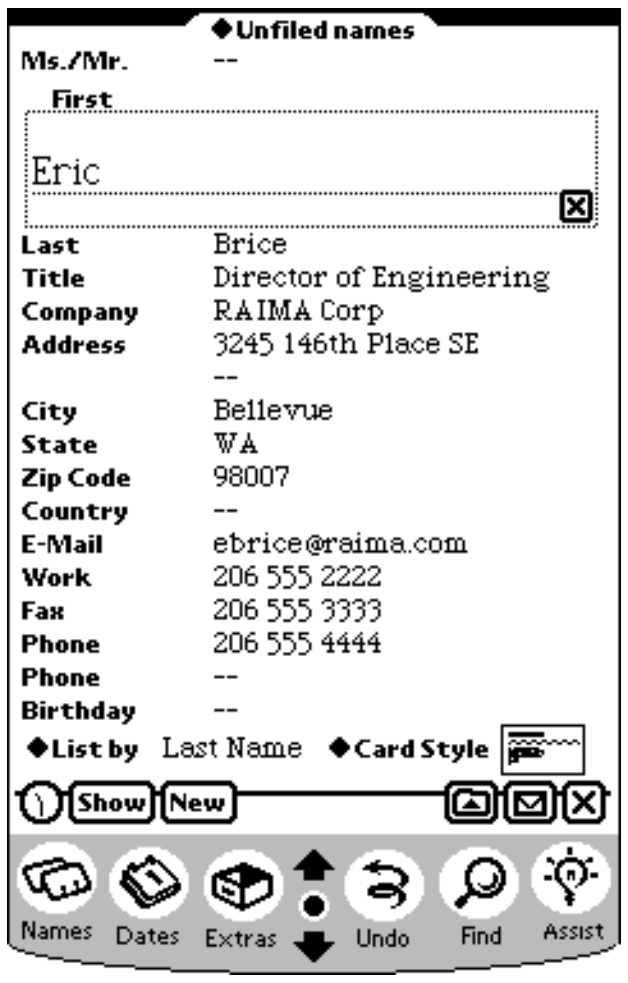

Figure 1: Names application included with all Newton computers depicted at one quarter life size of the Apple Newton MessagePad 100 used in the experiments. As the user taps on a field, it expands to ease writing. In this picture, the First Name field is expanded. The folder tab button at the top of the screen is for displaying names from one of eleven user-defined folders. From left to right, buttons at the bottom of the application screen are for showing the time and battery state (labeled with a clock face), changing the display of this name (labeled "Show"), adding a new name (labeled "New"), refiling this name (labeled with the file folder picture), printing/faxing/infrared beaming/mailing/duplicating/deleting this name (labeled with the envelope picture), and closing the application (labeled with a large "X"). Below these are universal buttons visible in all applications. From left to right, they provide access to the Names application, a calendar application, a storage place for all other applications, scrolling buttons, undo, find, and natural language recognition.

## 3. Names++

A Newton computer's built-in Names application includes one of the three components suggested by Section 1 to speed adding a new person's name. It recognizes handwriting, and

5. Throughout this paper references to typing refer to tapping on an on-screen soft keyboard.

the recognition dictionary can be expanded as needed. Names++ is an extended version of Names which we wrote to include the other two components.

### 3.1 Adaptive Menus

Names++ extends Names by adding an adaptive menu to 9 of Names’ 17 fields: 7 menus consisting of 4 recently entered values and 2 menus with 4 recently entered values prepended to fixed choices. If the word the user needs is in a menu, they can choose it rather than write it out. Figure 2 depicts Names++ with a menu open for the City field. The choices in the menu are the four most recently entered values for this specific field. (Each field has a separate menu.) This may be the most convenient when the user has a series of related names to add, perhaps for people from the same company or city. Of course, when the user adds four unusual values in a row, common choices are inadvertently dropped from the menu. A more sophisticated approach would list some number of the most recent values and some number of the most common; Names++ doesn’t explore this for the sake of simplicity and speed. Because these menus are adaptive, users will have to use linear search to examine choices and cannot rely on muscle-level memory of choice locations. If the menus include more than a few choices, the cost of this search will likely dominate any Fitts’ law effect.[6]

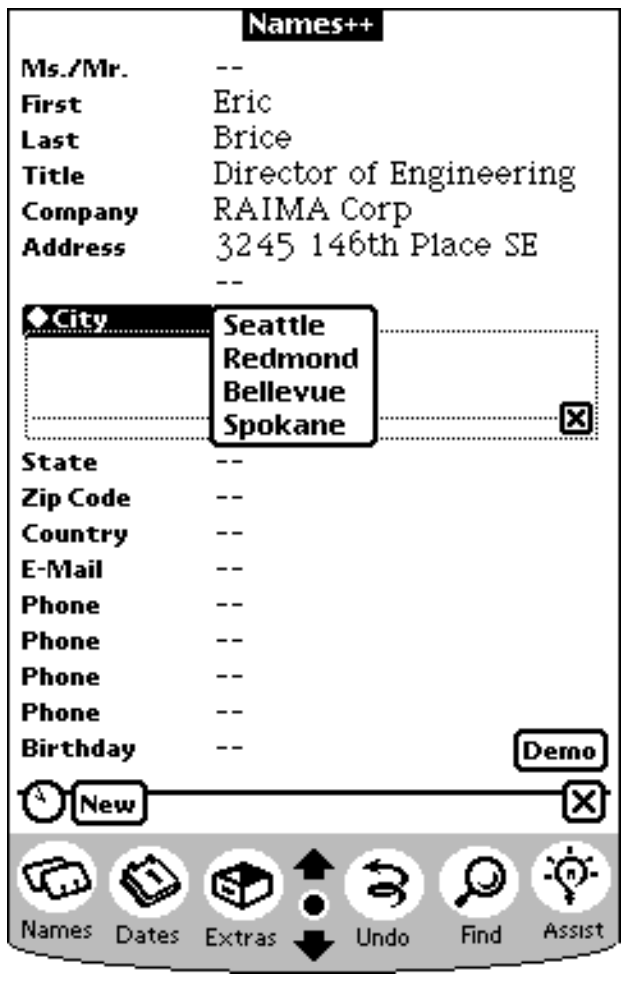

Figure 2: Names++ application. In this picture, the user has tapped on the word “City” and opened a menu of recently used city names. If the user chooses one of these cities, it will be copied into the City field for this name. Compare this to Figure 1. Note that Names++ includes all features of Names relevant to adding a new name.

Two of Names’ fields already had a menu. The Honorific field offered the user “Ms.”, “Mrs.”, “Mr.”, and “Dr.”; the Country field offered a menu of thirteen countries. For completeness, Names++ prepends the four most recent values to these fields’ menus. Technically these are *split* menus.[7] Mitchell and Shneiderman (1989) compared large statically ordered menus (unsplit) to those that also prepended most-recently-used choices

6. Fitts’ law states that the time to move a given distance $D$ to a target of width $W$ is proportional to the log of $D/W$.
7. Not to be confused with splitting menu choices across multiple menus in (Witten, Cleary, & Greenberg, 1984).

(split, exactly our condition). Static were faster than split menus on one task; there was no difference on another. Sears and Shneiderman (1994) later found evidence in favor of split menus including a 17–58% improvement in selection time compared to unsplit menus. They also compared alternative organizations for the split part and recommend limiting the number of split choices to four or less (which Names++ does) and sorting split choices by frequency (which Names++ approximates by most-recently-used). In the context of menu hierarchies, Snowberry, Parkinson, and Sisson (1985) found that adding items containing upcoming selections resulted in greater accuracy and faster search times. This result has not been confirmed (Kreigh, Pesot, and Halcomb, 1990) and may be as much an effect of preventing users from getting lost in menu hierarchies as of assisting them in making selections *per se*. We test adaptive (split) menus in Names++ to understand the relative contribution of such compared to other interfaces in a data entry task.

The four Phone Number fields have menus, but these give the user a way to categorize the phone number rather than enter the number itself. They are *category* menus (Norman, 1991). The phone menus include choices for “Phone”, “Home”, “Work”, “Fax”, “Car”, “Beeper”, “Mobile”, and “Other”. All phone fields have identical menus. Names++ does not modify them. No menus were provided for the First Name, Last Name, and Birthday fields. Section 5 describes which input fields should have menus.

To understand the computational space and time demands of adaptive menus, note that Names++ stores all menus in a single object in its own object database. The size of the object is linear in the number of fields with menus ($f$) and in the number of choices in each menu ($c$), or $fc$. If each menu were implemented as a circular queue, the time to update the object would be constant for each menu, or just $f$. Names++ uses a slightly slower array implementation for menus and takes $fc$ time. In practice this works out to slightly more than one half second for nine fields and four choices.

### 3.2 Predictive Fillin

Names++ also extends Names by automatically filling up to 11 empty fields in a new name with predicted values. It treats previous names as a case base (Kolodner, 1993) and copies information from a relevant case. Specifically, when the user adds a company to a new name that matches a previous name’s company, Names++ copies most of the address from the previous name into the new one. Values are copied verbatim from the two address lines and the City, State, Zip Code, and Country fields. The user ID from the electronic mail address is dropped before the e-mail address is copied into the new name. (The remaining components of the e-mail address are likely to be the same for people from the same company.) The last word of Phone Number values is dropped; only the area code and prefix are copied into the new name for a typical area code-prefix-extension phone number. If the user writes or chooses another value for Company, replacing the the value of that field, predictive fillin recopies dependent values from a previous name. Figure 3 illustrates the sequence of events from the user’s perspective.

Names++ behaves similarly when the user adds a city or state that matches a previous name, but it copies over less information than when a matching company is found. If a value copied by predictive fillin is incorrect, the user can write or choose the correct value manually. Table 1 summarizes which fields have menus and predictive fillin. The structure of predictive fillin is fixed in the design of Names++. Other work attempts to learn comparable structure from examples (e.g., Dent et al., 1992; Hermens & Schlimmer, 1994; Schlimmer &

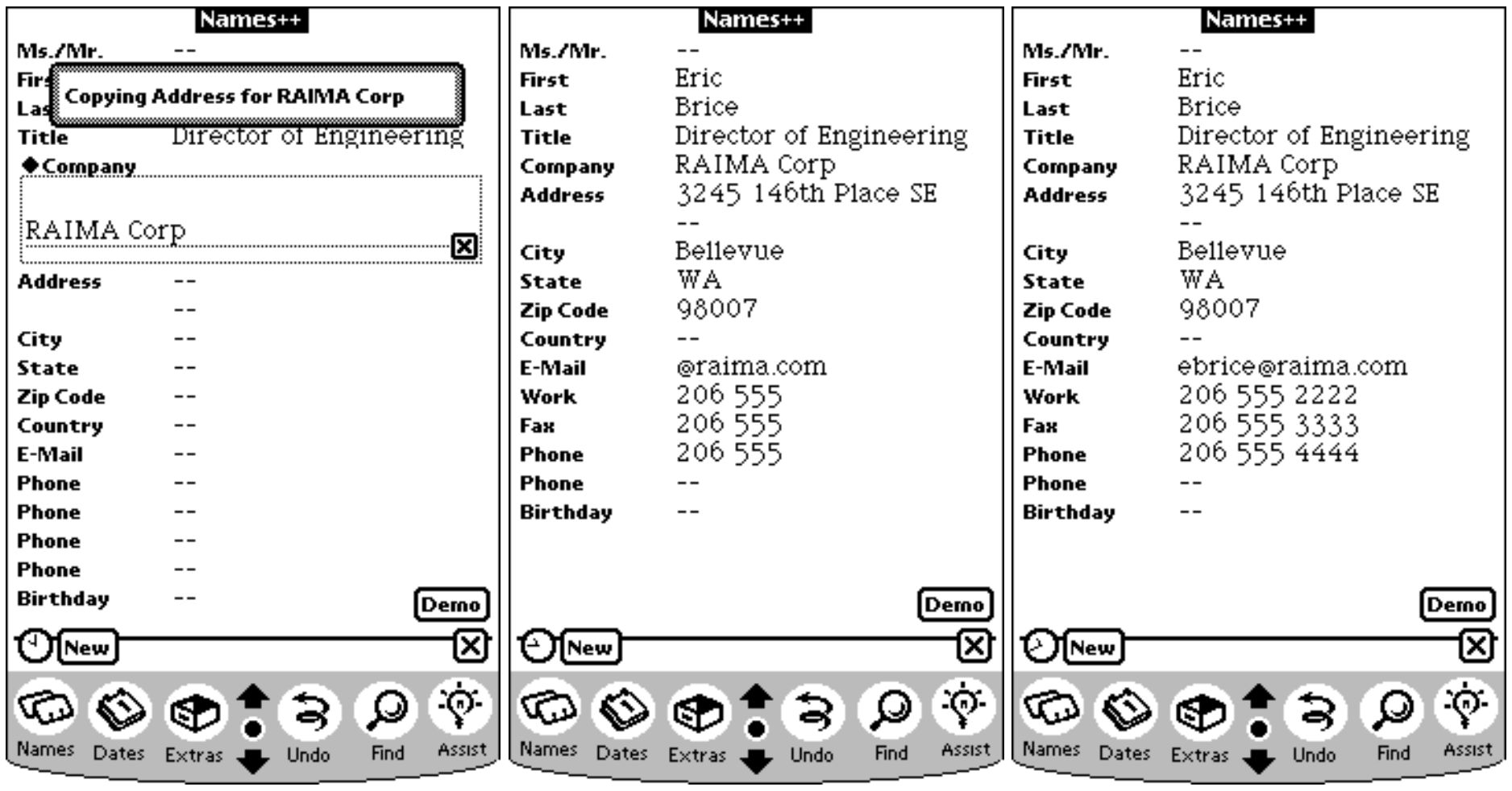

Figure 3: Names++ application after the user adds a company for a new name. In the left panel, the application finds a previous name with a matching company, displays a dialog, and fills remaining fields with predicted information copied from the previous name. The center panel shows how much information was filled. The right panel shows the completed name. In this example the user has written only four additional words to complete the name.

Hermens, 1993; Yoshida, 1994). Our goal is to determine whether the end result of such learning is worthwhile.

If more than one previous name matches the company, city, or state of the new name, Names++ fills fields with values from the most recent name. Values from the second-most recent occurrence of that name are added to menus. This gives the user a chance to select between alternate addresses for the same company or alternate zip codes for the same city.

In terms of computational requirements, Names++ needs no additional storage for predictive fillin; the object database of previously added names is reused as a case base. Matching a new name's company, city, or state to a previous name is implemented with a Newton primitive; an informal study depicted in Figure 4 indicates that Newton's proprietary algorithm appears to run in time linear in the number of names if no match is found and logarithmic if matches are found.

Names++ and its source code is in on-line Appendix A.

## 4. Experiments

We hypothesize that recognition, adaptive menus, and predictive fillin speed adding a new name. To find out if this is so and to what extent, we conducted an experiment in which subjects added names using different combinations of the three interface components.

### 4.1 Method

Five computer science students between the ages of 18 and 35 years of age participated as subjects in the experiments. Prior to the experiments they had used a Newton computer for at least six months and were familiar with Newton's handwriting recognition and QWERTY layout of Newton's on-screen keyboard.

The experiment used a within-subject design where each subject participated in each of six conditions summarized in Table 2. Conditions were designed to assess the contribution of each interface component separately and collectively. In the control, Typed condition, the subject types all values without using any of the components. In the Null condition, the subject writes all words using remedial recognition steps (to be described) and types words only if they are not recognizable. The subject does not add words to Newton's dictionary when asked and does not have the assistance of either adaptive menus or predictive fillin. The D condition extends Null by requiring the subject to add words to Newton's dictionary when asked. The AM condition extends Null by adding adaptive menus. The PF condition extends Null by adding predictive fillin. The All condition combines the extensions of D, AM, and PF.

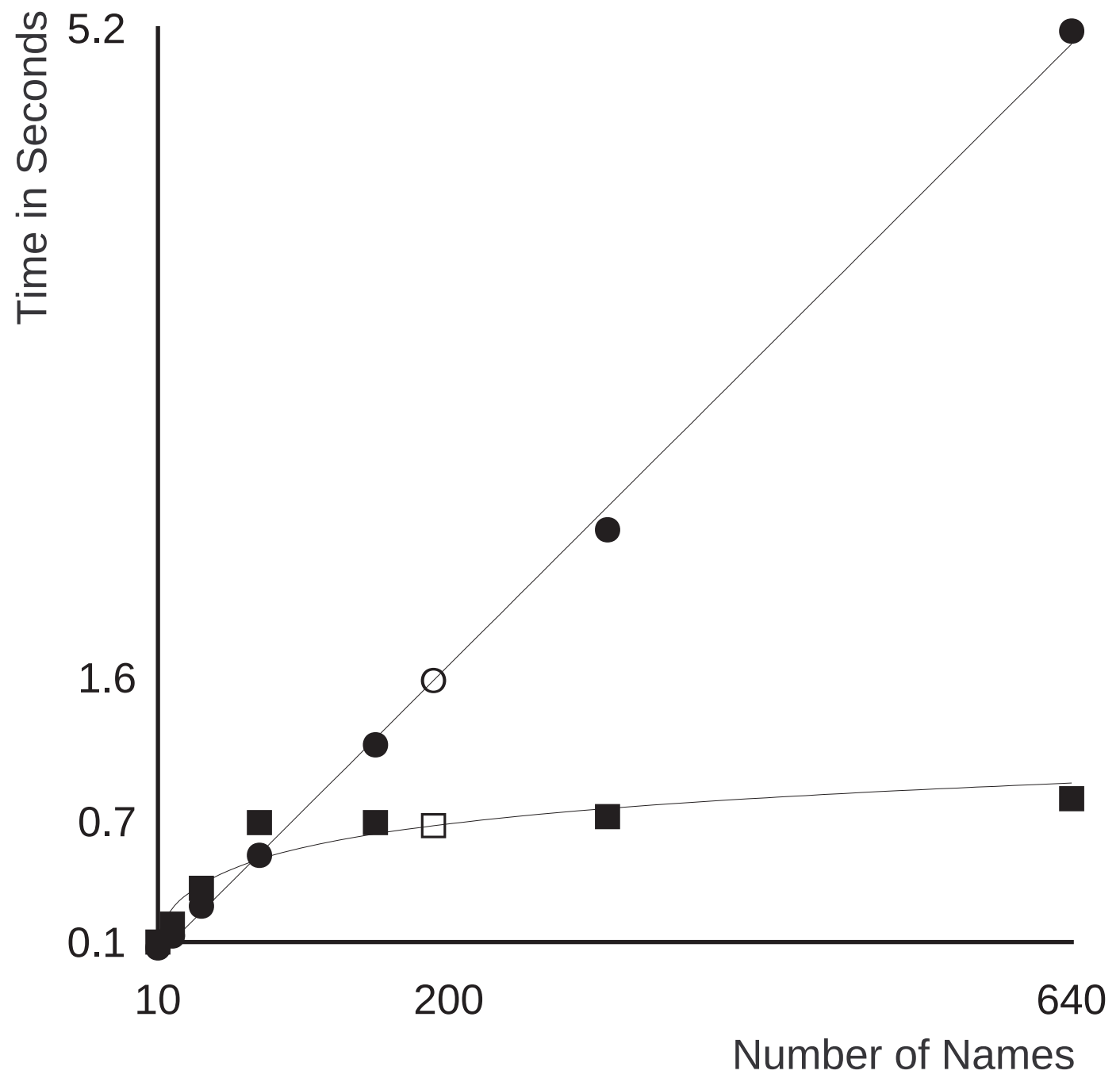

Figure 4: Time to find a matching name using Newton as a function of the number of names in the database if no match exists (circles and upper line) and if several matches exist (squares and lower line). Both axes are linear scale. The upper line is a linear fit; the lower line is logarithmic. For comparison to the experiment, interpolated values for 200 names are shown with open symbols.

We used a pair of Apple Newton MessagePad 100 computers (running Newton OS version 1.3) for the experiment and three versions of the Names++ application. One version has all interface components disabled and was used for the Typed, Null, and D conditions. A second version has adaptive menus and was used for AM. A third version has adaptive menus and predictive fillin and was used for PF and All.

A set of 448 name records for the experiments was donated by a development officer from Washington State University. Her job involves contacting alumni and others to solicit support for university programs. Almost all of the records include a first and last name, a full mailing address, and one to three phone numbers. Few include an honorific, country, or e-mail address. Informal tests indicated that the MessagePads could hold about 250 names after Names++ was installed, so we selected a random set of 200 of these records.

| Field | Menu Choices | | Predictive Fillin by | | | Notes |
|---|---|---|---|---|---|---|
| | Built-in | Adaptive | Company | City | State | |
| **Honorific** | 4 | 4 | | | | “Ms.”, “Mrs.”, “Mr.”, “Dr.” |
| **First Name** | | | | | | |
| **Last Name** | | | | | | |
| **Title** | | 4 | | | | |
| **Company** | | 4 | | | | |
| **Address (1)** | | 4 | Yes | | | |
| **Address (2)** | | | Yes | | | No label to tap for menu. |
| **City** | | 4 | Yes | | | |
| **State** | | 4 | Yes | Yes | | |
| **Zip Code** | | 4 | Yes | Yes | | |
| **Country** | 13 | 4 | Yes | Yes | Yes | |
| **E-Mail** | | 4 | Yes | | | User ID is removed. |
| **Phone 1** | 8 | | Area Code and Prefix | Area Code | | Category menu is used to select type of phone number rather than the phone number itself. Choices include “Phone”, “Home”, “Work”, “Fax”, “Car”, “Beeper”, “Mobile”, and “Other”. |
| **Phone 2** | 8 | | | | | |
| **Phone 3** | 8 | | | | | |
| **Phone 4** | 8 | | | | | |
| **Birthdate** | | | | | | |

Table 1: Name++ fields with adaptive menus and predictive fillin.

| Condition | Writing | Add to Dictionary | Adaptive Menus | Predictive Fillin |
|---|---|---|---|---|
| **Typed** | | | | |
| **Null** | Yes | | | |
| **D** | Yes | Yes | | |
| **AM** | Yes | | Yes | |
| **PF** | Yes | | | Yes |
| **All** | Yes | Yes | Yes | Yes |

Table 2: Experimental conditions, one row per condition. Columns indicate which user interface components were used. Blank cells represent “No.”

To simulate a worst case for recognition, adaptive menus, and predictive fillin, we chose 5 names (listed below) from the residual 248 such that each name’s company was not in the preload set of 200 names. (To preserve anonymity here, first and last names are swapped and phone numbers replaced with artificial values. Actual first and last name pairs and phone numbers were used in the experiment.)

Robert Anderson
Account Marketing Rep
IBM
W 201 N River Drive
Spokane, WA 99201
509 555 0000
509 555 1111

Eric Brice
Director of Engineering
RAIMA Corp
3245 146th Place SE
Bellevue, WA 98007
206 555 2222
206 555 3333
205 555 4444

Mike Carlson
VP Engineering & Estimating
General Construction
2111 N Northgate Way
Suite 305
Seattle, WA 98133
206 555 5555
206 555 6666

Peter Friedman
President
NOVA Information Systems
12277 134th Court NE
Suite 203
Redmond, WA 98052
206 555 7777

Thomas Leland
Staffing Manager
Aldus Corporation
411 First Ave South
Seattle WA 98104 2871
206 555 8888
206 555 9999

To score how words in names were entered and the total time, we used the sheet in Figure 5. Fictitious data corresponding to a subject's entering of the second name in the All condition is also depicted.

| | | | | | | | | | | |
|---|---|---|---|---|---|---|---|---|---|---|
| | Subject | | | ROMs | MP 100 | | | | | |
| | Date | 13 Apr 96 | | | | | | | | |
| X | Adding to dictionary | | | X | First Time This Name Entered | | | | | |
| X | Using adaptive menus | | | | Second Time This Name Entered | | | | | |
| X | Using predictive fillin | | | | | | | | | |
| | | Value Entered | Recognized Correctly | In Recog. Menu 1 | Recognized Letter/Letter | In Recog. Menu 2 | Had to be Typed In | Asked to Add to Dictionary | In Adapt. Menu | Predictive Fillin |
| | Honorific | | | | | | | | | |
| | First Name | Eric | 1 | | | | | | | |
| | Last Name | Brice | | | 1 | | | | | |
| | Title | Director of Engineering | 1 2 3 | | | | | | | |
| | Company | RAIMA Corp | 2 | | | | 1 | 1 | | |
| | Address | 3245 146th Place SE | 1 3 | | | | 2 4 | 2 4 | | |
| | Address 2 | | | | | | | | | |
| | City | Bellevue | | | | | 1 | 1 | | |
| | State | WA | | | | | | | 1 | |
| | Zip Code | 98007 | 1 | | | | | | | |
| | Country | | | | | | | | | |
| | E-Mail | | | | | | | | | |
| | Phone | 206 555 2222 | 1 2 3 | | | | | | | |
| | Phone | 206 555 3333 | 1 2 3 | | | | | | | |
| | Phone | 206 555 4444 | 1 2 3 | | | | | | | |
| | Phone | | | | | | | | | |
| | Birthday | | | | | | | | | |
| | Duration | 3:51 | | | | | | | | |

Figure 5: Scoring sheet used each time a name was added. A "1" in the center to right columns indicates how the first word of a field value was entered using recognition (cf. Figure 6), an adaptive menu (cf. Figure 2), or predictive fillin (cf. Figure 3). A "2" indicates the second word, and so on. The highest digit in a row corresponds to the number of words in that field's value.

To facilitate setting up each condition, we constructed backup images of MessagePads correctly configured for each of the six conditions. For all images, the 200 names and the appropriate version of Names++ was installed. To the images for D and All, we added all First, Last, and Company names to its dictionary using a built-in feature of Newton. To initialize the adaptive menus in the images for AM and All, we used a special purpose application. Prior to each use the MessagePads were completely erased and then restored from the backup image appropriate to the condition to be tested.

The task of the subject was to enter each of the five names twice in each of the six conditions. The first time a name is entered in a condition simulates a worst-case scenario; the second time, a best.

## 4.2 Procedure

Subjects were given a listing of one of the five names and a MessagePad initialized for one of the six experimental conditions. Each subject entered each name in each condition; name/condition pairs were randomly ordered for each subject to counteract subject learning effects. They were instructed to enter names quickly. Subjects made few mistakes. They were instructed to correct these before finishing. Times reported below include time to correct mistakes.

Subjects were given a precise script to follow when entering a name. This was done partially to bias results against the hypotheses and partially to minimize individual variation. Specifically, the subject was instructed to enter values for each field in order, from top to bottom, completing one before going to the next (cf. Figure 1). In conditions involving handwriting, if a word was not correctly recognized, the subject was to check the menu of alternate recognitions (depicted in the left panel of Figure 6). If the intended word was not in this list, they were to select "Try letters" which attempts recognition without the dictionary. If the result of this was not correct, they were to check a second menu of alternative recognitions (depicted in the center panel of Figure 6). If the intended word was not in this second menu, they were to tap the button with the keyboard picture, type in the word using the on-screen keyboard, and close the keyboard. If this word was not already part of the dictionary, Newton asked if they would like to add it (depicted in the right panel of Figure 6). Note that for each of the recognition menus, the original handwriting is shown near the bottom. The first choice is Newton's best guess, and the second choice is its best guess with different capitalization. The subject was instructed to ensure that words were correctly capitalized.

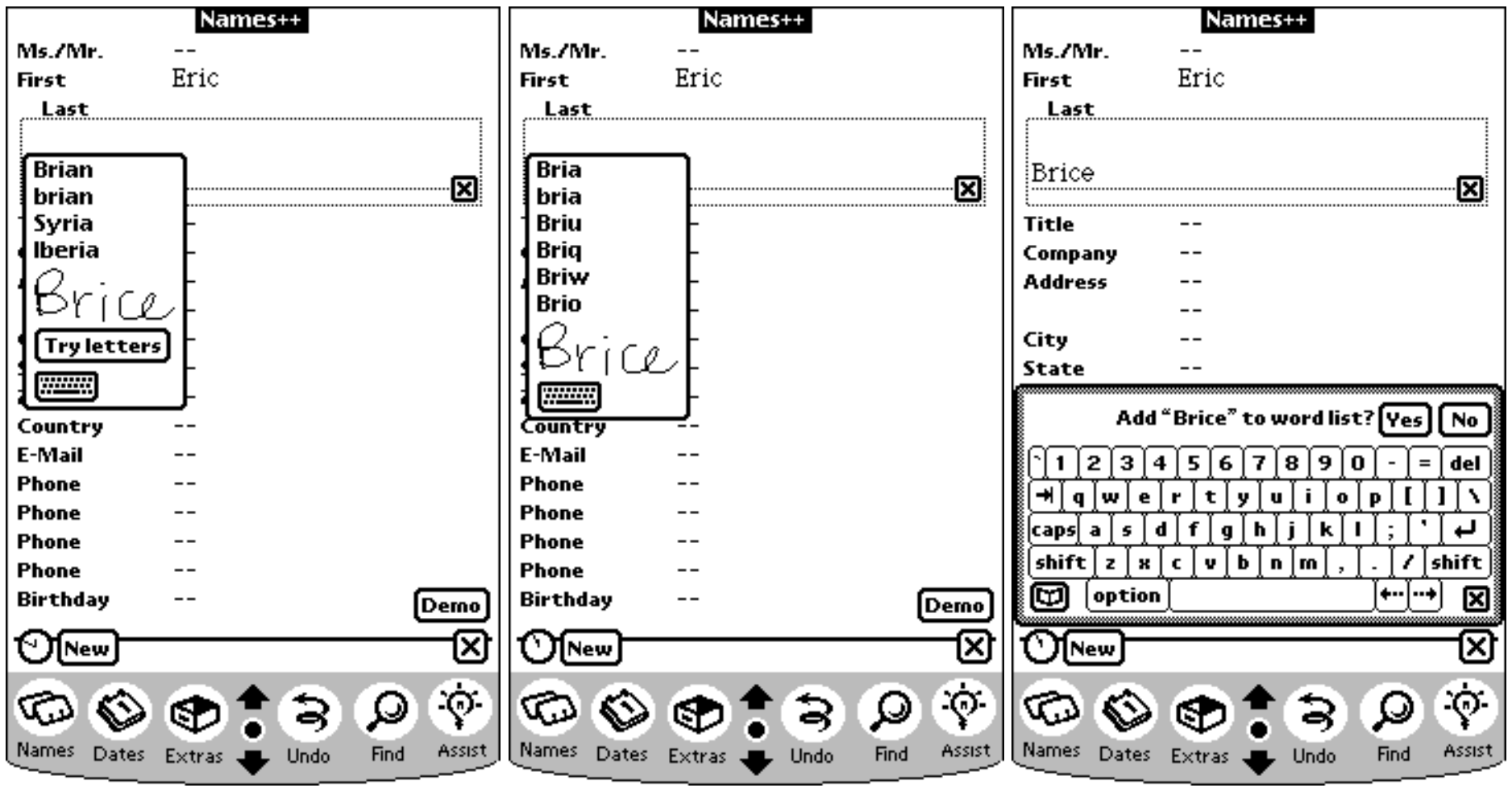

Figure 6: Remedial steps when a handwritten word is not correctly recognized. In this example, the subject wrote "Brice" which was misrecognized as "Brian". When the subject double-taps on the word, a menu of alternative recognitions appears (left panel). If none of these are correct, the subject requests recognition without the dictionary (or letter by letter). Another double-tap on the word generates a second menu of alternatives (middle panel). If none of these are correct, the subject entered the word by tapping on the buttons of an on-screen keyboard (right panel).

For Typed, the subject was instructed to enter all data using Newton's on-screen soft keyboard. For Null, the subject was to enter all data by handwriting. For D and All, the subject was instructed to add any words to Newton's dictionary if asked. For AM and All, the subject was instructed to check a field's menu (if there was one) before writing any data. No special instructions were required for PF beyond the default of not adding words to the dictionary.

A stopwatch was started when a subject tapped the "New" button and stopped when the last field value had been correctly entered. Choosing a manual timing method simplified development of the experimental software. The method by which each word of each field was entered was recorded on a scoring sheet as indicated in Figure 5.

The experiment took between three and five hours for each subject and was spread over two or more sessions of approximately two hours within the same week. Subjects took short breaks after adding each name to minimize fatigue.

After each subject completed the experiment, they were asked to rank their favorite methods for entering names from most to least.

### 4.3 Results

Table 3 summarizes the median and standard deviation of the subjects' time to enter a name in the six conditions. Times include all user input, predictive fillin computation, and time to correct errors (if any). The first row reports time to add a novel name, our simulation of a worst case. The second row reports time to repeat the name, our simulation of a best case. An ANOVA reveals a significant main effect for condition $F(5, 21.07) < 0.001$. The interaction of number of times a name is entered and condition is also significant $F(5, 19.61) < 0.001$. Comparing worst cases across conditions, a post-hoc multiple comparisons test using Tukey's HSD indicates that only Typed is significantly different (faster) than other conditions. (All $p < 0.05$.) Comparing worst to best cases within the same condition, D, AM, PF, and All are significantly faster. Comparing best cases across conditions, Typed, AM, and PF are significantly faster than D and Null; All is significantly faster than Typed, PF, D, and Null. No other pairwise comparisons are significant.

| | Typed | Null | D | AM | PF | All |
|---|---|---|---|---|---|---|
| **Worst** | 2.72 (0.86) | 4.25 (1.31) | 4.50 (1.45) | 4.32 (1.70) | 4.07 (1.26) | 4.15 (1.13) |
| **Best** | 2.52 (0.60) | 3.65 (1.24) | 3.30 (1.09) | 1.37 (0.51) | 2.02 (0.45) | 1.08 (0.24) |

Table 3: Median time in minutes to add a new name over five names and five subjects (25 samples per cell, standard deviation in parentheses). Columns list six experimental conditions.

The difference within D, AM, and PF across worst and best cases confirms our hypothesis that these interfaces can speed entering names, by 29%, 210%, 110% compared to Null, respectively. We were surprised to find that predictive fillin was not as fast as adaptive menus (though the difference is not statistically significant). When designing a data entry system one might be tempted to implement just adaptive menus given their algorithmic simplicity, especially compared to sophisticated methods in machine learning that have been proposed for predictive fillin. However, the latter do not suffer from recency effects imposed by the limited size of adaptive menus; when entering new data related to some in the distant past, predictive fillin would have little difficulty providing assistance where adaptive menus could not. Adaptive menus could be further refined to use a frequency or frequency-recency combination, but the performance of All suggests implementing both adaptive menus and predictive fillin. Combined with adding words to a dictionary, they can speed entering names by 294%. In practical terms, these interfaces could make entering a name into an electronic organizer *faster* than writing it down on paper and certainly fast enough to capture the information during a phone conversation.

Prior work confirms the difference between the Typed and D conditions. Ward and Blesser (1986) state that normal writing speed is rarely greater than 69 characters per minute (cpm) for a single line of text. Using the fact that the mean number of characters per name in our experiment is 98.2, our subjects achieved 30 cpm. MacKenzie, Nonnecke, Riddersma, McQueen, and Meltz (1994) compare four interfaces for entering numeric and text data on pen-based computers, including hand printing and using an on-screen keyboard. (The other two interfaces were experimental gesture-based techniques for entering single characters.) For numeric entry conditions, they found that the on-screen keyboard was 30 words per minute (wpm) with 1.2% error whereas hand printing was 18.5 wpm with 10.4% error. For text entry conditions, the keyboard was 23 wpm with 1.1% error whereas hand printing was 16 wpm with 8.1% error. Using the fact that the mean number of words per name in our experiment is 20.8, our subjects achieved 8.3 wpm typing and 6.3 wpm handwriting for mixed numeric/text input. The key point of comparison is that both their and our studies found that using a stylus to tap an on-screen keyboard is faster than handwriting or printing. Differences in speed between these studies and ours is likely a result of differences between experimental procedures (theirs versus ours): single versus multiple field fillin, copying information from memory or screen versus paper, and block or comb-type (letter) versus open (word) interface.

Figure 7 presents Box plot summaries of the time data. Of interest is reduction in variance of time by adaptive menus and predictive fillin in the best case (right plot). Differences between individual performance is reduced by these interface components.

The left half of Table 4 lists the recognition accuracy for each field over all conditions, subjects, and names. The first row indicates that 94% of the first names written were correctly recognized immediately. By checking the first menu of alternate recognitions, that accuracy rises to 95%. Similarly, the second row indicates that 59% of all second names written were correctly recognized immediately. This rate rose to 74% when letter-by-letter recognition was invoked and again to 79% by checking the second menu of alternate recognitions. Phone numbers enjoyed the second highest recognition rate below first names.

For reference, Cesar and Shinghal (1990) report over 90% recognition rate on hand printed, Canadian postal codes which are {letter, digit, letter, space, digit, letter, digit}. This is comparable to our observed rates for first names, second address lines, and phone numbers.

The right half of Table 4 lists the percentage of words entered using typing, adaptive menus, or predictive fillin by field over all conditions, subjects, and names. The first row indicates that 5% of first names were typed. The row for State indicates that 32% of state names were typed, 20% were chosen from an adaptive menu, and 39% were predictively filled in. (Note that the numbers in each row do not total to 100% because the left half of the table lists percentages for words that were *written* while the right half lists percentages of *all* words.)

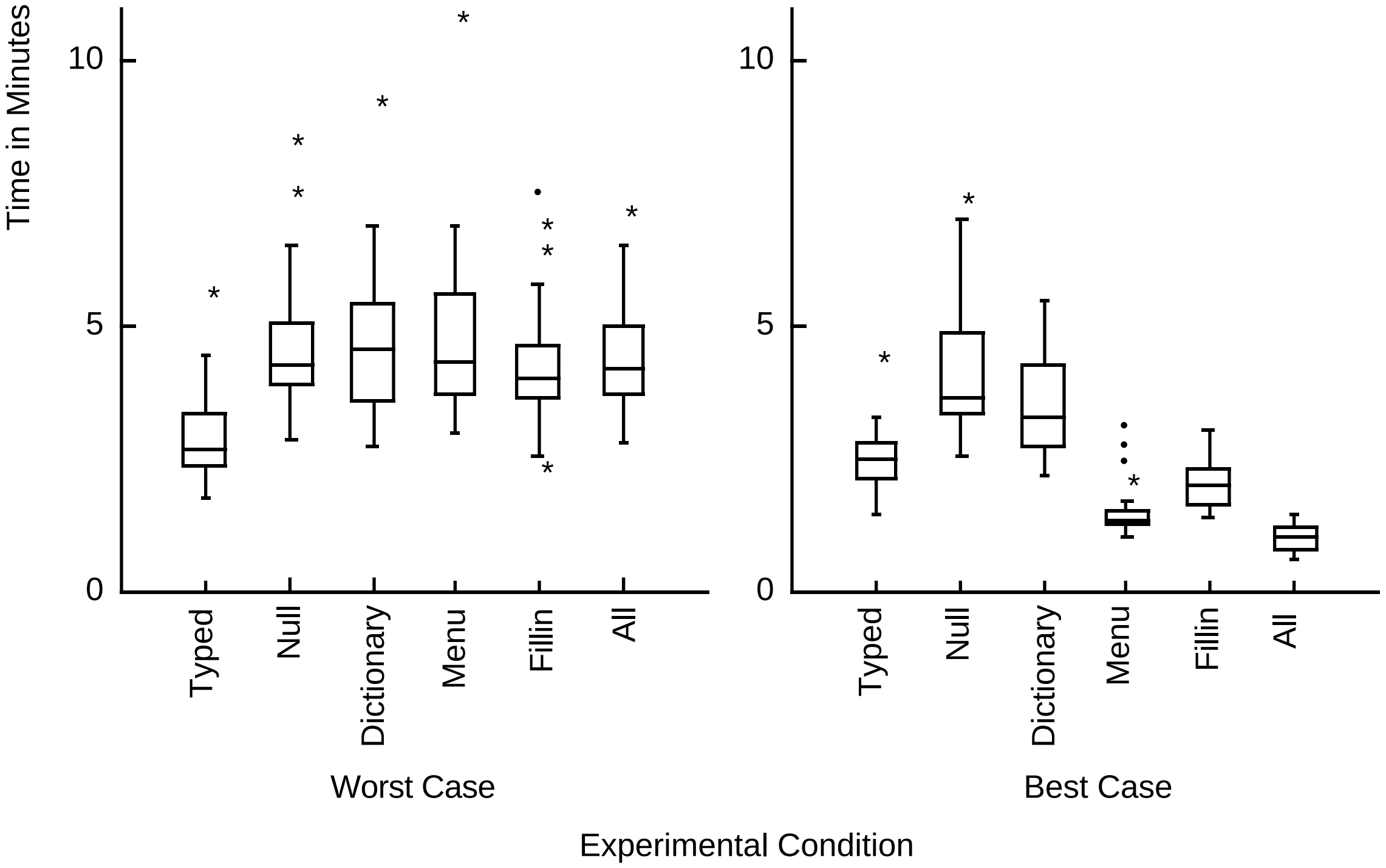

Figure 7: Box plots of time to enter a name by condition in the worst and best cases. Each box summarizes 25 values. Values outside the inner "fences" are plotted with asterisks. Values outside the outer "fences" are plotted with circles (Wilkinson, Hill, Vang, 1992).

Combining the left and right halves of Table 4 reveal that many of the difficult-to-recognize fields have considerable assistance from adaptive menus and predictive fillin. This accentuates the speed improvements by providing help where it is most needed. Figure 8 depicts the relationship between the fields, their recognition accuracy, and which have adaptive menus or predictive fillin. Several fields have near perfect recognition accuracy; they can be recognized without resorting to typing. For instance, numeric fields are easier to recognize; the Phone Number fields were recognized at nearly 90% even though the area code, prefix, and suffix varied from name to name. The First and Last name fields also had high recognition accuracy. All of the first names were in the built-in dictionary. All but two of the last names were, and the others were often recognized letter by letter. Recognition was poorer in the Company and Address fields. Words in full capitals (e.g., "RAIMA") and words with a combination of numbers and letters (e.g., "146th") were difficult to recognize. The low recognition accuracy of the State field is apparently due to an oversight in Newton's dictionary. "WA" is not included but many other two-letter abbreviations for US states are. To compensate for low accuracy, Names++ includes an adaptive menu and/or predictive fillin for each of the difficult-to-recognize fields.

Table 5 summarizes subjects' preference for condition to enter a name. It lists frequency of ranking over the five subjects. Subjects partitioned conditions into non-overlapping groups of (Typed, Null), (D, AM, PF), and (ALL). (The authors know of no suitable statistic for asserting these differences.) These results contradict those of MacKenzie et al. (1994) who found that subjects preferred typing to handwriting, mildly for text entry and more strongly for numeric entry. They restricted hand printing input to block or comb-type interface; this unnaturalness may account for some of the dispreference toward handwriting. Writing with a stylus does have its advantages. As Meyer (1995) points out, keyboards are faster for linear text entry, but a pen input device can be more natural, can handle text and graphic input, and can jump quickly from point to point. Writing with a pen also supports “heads up” writing, allowing the user to visually attend to other aspects of the task at hand. Typing with an on-screen keyboard requires heads down entry.

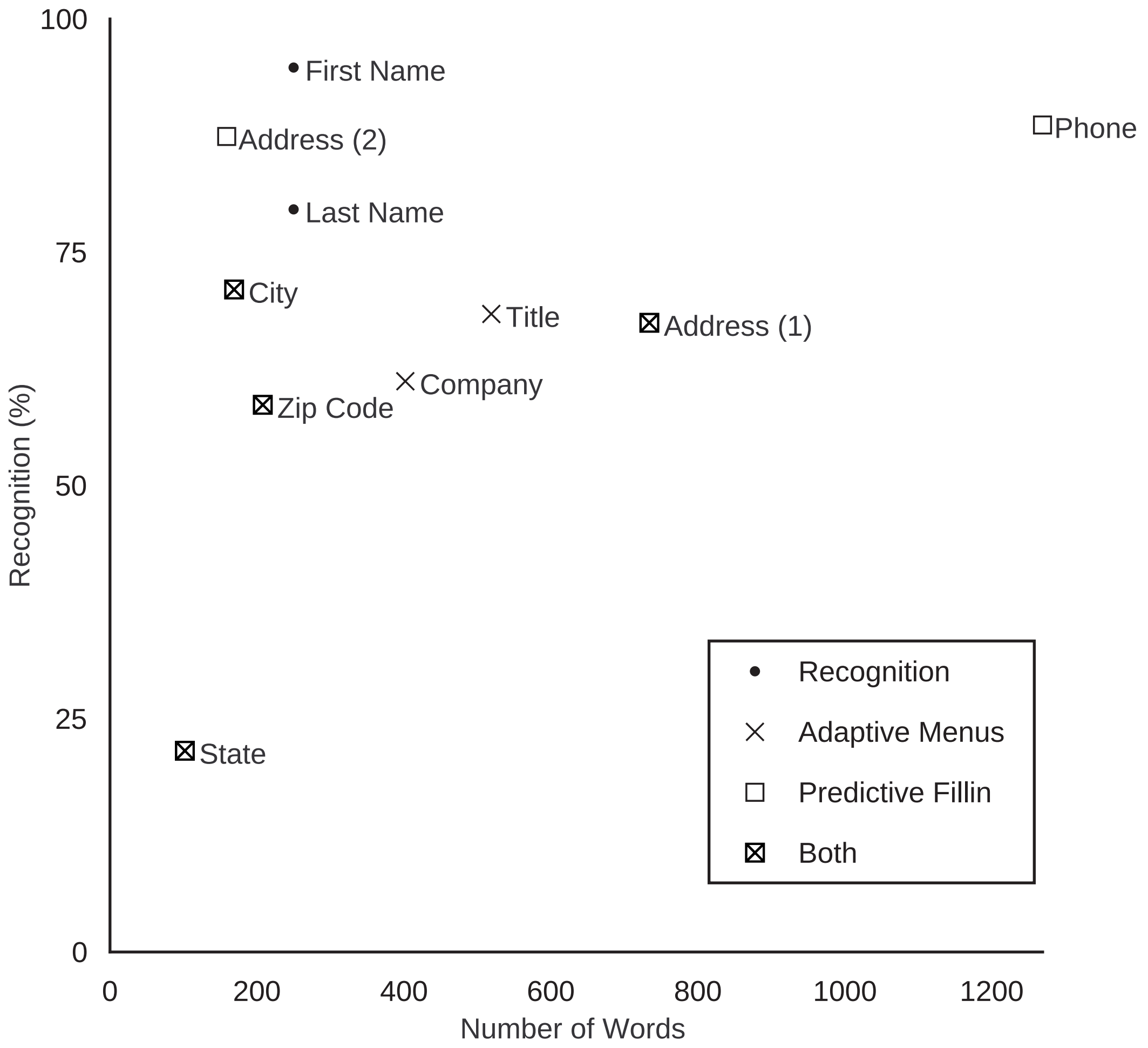

Figure 8: Recognition rate as a function of the number of total words entered in all conditions by all subjects for all names. Fields with adaptive menus or predictive fillin (or both) are marked. Note that every field with less than 75% accuracy has either an adaptive menu or predictive fillin (or both).

One subject experimented with Names++ outside the experimental setting and offered a number of observations. First, the adaptive menus were too short, and sometimes menus would be useless no matter how long they were. She wished that the City and Company field’s menus were longer (especially City). It was frustrating to have one of the common city names for a large metropolitan region bumped from the short list. In contrast, the Title field’s menu was rarely useful, and she did not see the point of maintaining it. The principles to be outlined in Section 5 suggest similar revisions.

| Field | Cumulative Recognition Accuracy | | | | Percent Words Entered | | |
|---|---|---|---|---|---|---|---|
| | Correct | 1st Menu | Letter by Letter | 2nd Menu | Typed | Adaptive Menu | Predictive Fillin |
| **First Name** | 94 | 95 | 95 | 95 | 5 | | |
| **Second Name** | 59 | 59 | 74 | 80 | 21 | | |
| **Title** | 52 | 62 | 66 | 68 | 26 | 20 | |
| **Company** | 42 | 49 | 59 | 61 | 31 | 20 | |
| **Address** | 48 | 60 | 62 | 67 | 23 | 10 | 20 |
| **Address 2** | 81 | 85 | 87 | 87 | 10 | | 20 |
| **City** | 62 | 62 | 67 | 71 | 19 | 12 | 20 |
| **State** | 22 | 22 | 22 | 22 | 32 | 20 | 39 |
| **Zip** | 51 | 52 | 58 | 59 | 29 | 10 | 20 |
| **Phone** | 86 | 89 | 89 | 89 | 10 | | 15 |

Table 4: The left columns list cumulative recognition accuracy by field over all words that were written in all conditions, names and subjects. The right columns list percentage of all words by field entered by typing, adaptive menus, and predictive fillin. 5190 values total. Blank cells represent 0.

| Condition | 1st | 2nd | 3rd | 4th | 5th | 6th |
|---|---|---|---|---|---|---|
| **Typed** | | | | | 1 | 4 |
| **Null** | | | | | 4 | 1 |
| **D** | | | 1 | 4 | | |
| **AM** | | 2 | 2 | 1 | | |
| **PF** | | 3 | 2 | | | |
| **All** | 5 | | | | | |

Table 5: Subjects' frequency of ranking of preference for different conditions as a means to enter a name. 30 values total. Blanks cells represent 0.

Second, she found the predictive fillin helpful. Sometimes it filled when she didn't expect it to. She also noted that because predictive fillin copies over many fields, it encourages the user to add a more complete name. This may be an advantage in a harried setting.

## 5. Design Recommendations

Given these experimental results, how should we configure handwriting recognition, adaptive menus, and predictive fillin in another application (or a redesigned Names++)? For handwritten input, recognition should use dictionaries specific to each type of field: numbers only for numeric fields, and lists of domain terms for text fields.

### 5.1 Adaptive Menus

For adaptive menus, add a menu to any field that might have repeated values. If we accidentally added an adaptive menu to a field that never had the same value twice, our mistake would be harmless. The user would surely notice that the choices were useless and avoid checking the menu. When the menu was appropriate, the user would save time by choosing common values from it.

How long should each menu be? Long enough to include the most common values but short enough to be checked quickly. To make sure the menu is long enough, study how often a field's values repeat. For Names++, Figures 9a and 9b depict a frequency histogram for the

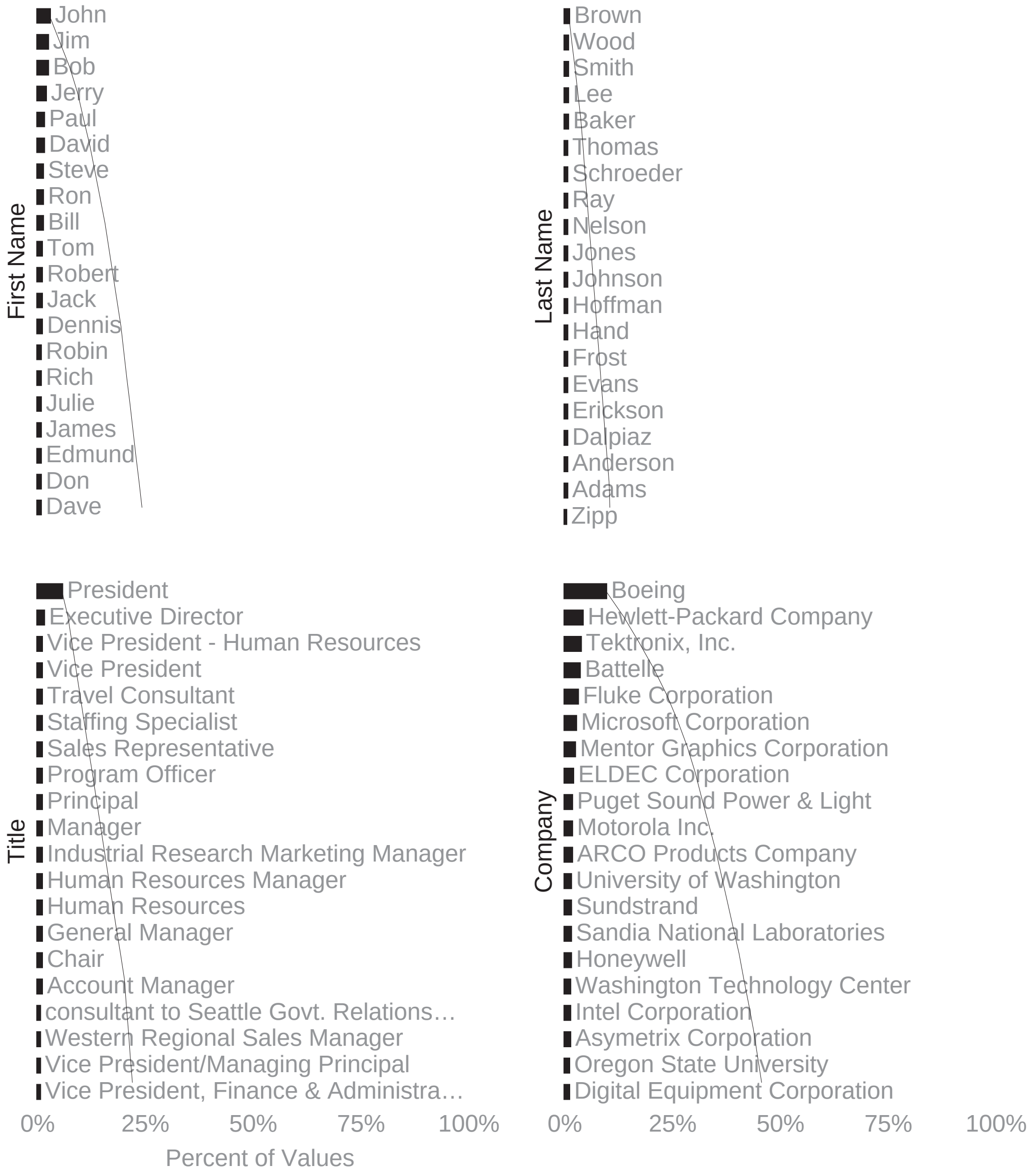

Figure 9a: Frequency of values in the First Name, Last Name, Title, and Company fields for the 448 names used in Section 4. Each plot is a histogram of the 20 most common values. Dark lines indicate what percent of values could be chosen by different sized menus. If the menu includes choices from the top down to its vertical position, it would allow the user to choose the percentage of field values indicated by its horizontal position.

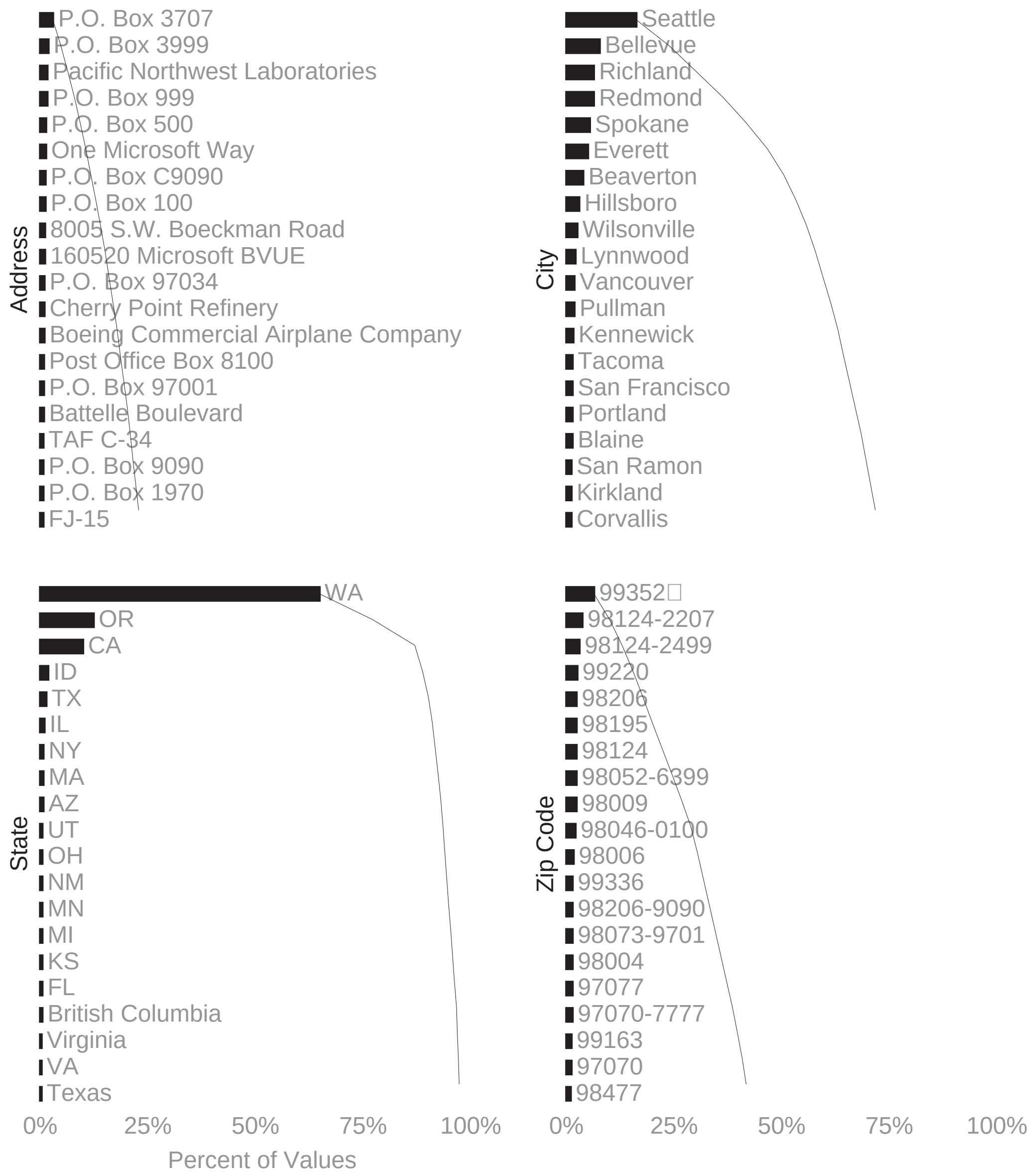

Figure 9b: Frequency of values in Address, City, State, and Zip Code fields for the 448 names used in Section 4.

20 most common values for 8 fields drawn from the 448 name records used in the experiments. Overlaid on each plot is a line indicating what percent of field values could be chosen from a particular size menu. For instance, for the First Name field in Figure 9a, the histogram is almost flat. A menu including only “John” would allow the user to choose a value for that field less than 5% of the time. If a menu included all 20 of the first names shown, the user could choose a value 25% of the time. This field should not have a menu because if it were long enough to include the most common values it would take too long to check. (Also, the Newton computer we used in Section 4 limits menus to 23 choices because of its screen size.) In contrast, for the Company field in Figure 9a, a menu including only "Boeing" would allow the user to choose a value more than 10% of the time. If it included the 20 values shown, the user could choose a value 50% of the time.

Studying the histograms and aiming for menus that include 50% of the field's values, we might re-engineer Names++ to have a menus of size 20 for the Company Field, size 10 for the City field, and size 5 for the State field. Other fields have very flat histograms and would need large menus to include a high percentage of field values. Recall that Section 4 reports one subject's frustration with the Title field. Only "President" seems to be repeated for this field in the 448 names we used.

### 5.2 Predictive fillin

Set up predictive fillin for any field that is functionally dependent (Ullman, 1988) on another. A functional dependency is related to the artificial intelligence idea of a determination (Russell, 1989). Intuitively, one field R, for *range*, functionally depends on another field D, for *domain*, if, given a value for D, we can compute a unique value for R. If predictive fillin can find a previous entry with the same value for D as the new entry, it copies over the previous entry's value for R into the new entry. In Names++, the Company field is the domain and the Address field is the range of a functional dependency.

Predictive fillin for all and only functionally dependent fields is probably too strict a strategy. Some functional dependencies are not useful for predictive fillin because their domain values are unique in the database. When this is so, predictive fillin cannot find a previously matching entry and cannot copy over relevant information. For instance, a US citizen's address is functionally dependent on their Social Security number. In an application like Names++ we don't expect to see the same Social Security number twice, so predictive fillin would never have the opportunity to help the user by filling the address. Functional dependencies with repeated domain values in the database, or *dense functional dependencies,* should be used to set up predictive fillin.

Conversely, some non-functional dependencies may be close enough to functional to be useful for predictive fillin. Technically, a dependency is not functional unless only one value in the range can be computed for every value in the domain. If only a few values in the range were computed for most values in the domain, the dependency might still be useful (Raju & Majumdar, 1988, Russell, 1989, Ziarko, 1992). For instance, most companies have a single office and address, but some may have more than one. It is still quite useful to fill address fields when Names++ finds a previous name with a matching Company field. Other user interface strategies can compensate for the other possible range values when they arise; for instance, Names++ puts alternate addresses into the Address field's adaptive menu. Therefore, dense dependencies that are functional or nearly so, or *dense approximately-functional dependencies*, should be used to set up predictive fillin.

To determine which dense approximately-functional dependencies hold for a new application area, it may be necessary to repeat the type of empirical domain analysis described above for adaptive menus. For Names++, we used common sense knowledge about people, companies, and addresses to set up predictive fillin. Recall that our goal is to discover if the end result of automatic learning is worthwhile (e.g., Dent et al., 1992; Hermens & Schlimmer, 1994; Schlimmer & Hermens, 1993; Yoshida, 1994). We recommend considering each field as a number of logical components because dependencies may exist between parts rather than whole fields. For instance, each person in a company may share a common telephone number area code and prefix, but they are likely to have different extensions. By predictively filling in all but the last component of a phone number, Names++ fills as much as it can without adding poor quality information.

## 6. Related Work

Though interested in different tasks, other researchers have studied using intelligent user interfaces to speed information capture. For instance, Hermens and Schlimmer (1994) built an electronic form filler that tried to provide default values for every field on the form. Each field of the form had a decision tree to calculate a default value. Like Names++, the calculations used previously entered information to generate defaults and predictively fill in fields. Unlike Names++, the calculations themselves were constructed at run-time using a machine learning method. (Names++ does not alter predictive fillin at run-time. cf. Table 1.) They field tested their system with a single electronic form filled out several hundred times over an eight month period. They report an 87% reduction in keystrokes; loosely translating this into a speedup yields 669% speedup or approximately 3 times the 210% speedup we observed for entering a name.

Studying text prediction without field boundaries, Pomerleau (1995) built a typing completion aid. Without relying on note-taking properties, his system predicts a completion for the current word being typed (presumably in an editor). A connectionist network estimates the probability of a number of possible completions for the current word; the most likely, over some threshold, is offered to the user. Pomerleau tested his system with a pair of subjects over a two-week period and reports an increase in typing speed of 2% for English text and 13–18% for computer program code. This modest gain may be due to inefficiencies in the learning method, to lack of redundancy in the task, or to limitations in the user interface itself.

As a complement to earlier research, this paper reports the individual and collective accuracy of three user interface components. It reports user task time showing that the components significantly improve efficiency. This paper also clarifies an issue confounded in earlier work. If a learning interface is less effective than expected, is this due to an inherent limitation in the interface itself, or does its learning method perform inadequately? To answer the second question, other work compares two or more learning methods. In this paper, we hand-built the predictive fillin structures (cf. Table 1) and were able to assess the quality of the predictive fillin interface directly.

## 7. Conclusion

This paper makes two main contributions. First, it presents a study of the impact of three user interface components on the time to enter information into a computer: handwriting recognition, adaptive menus, and predictive fillin. Handwriting recognition is slower than typing but is preferred by users. Advances in handwriting recognition may make it faster, but recognition would still be much slower than choosing a value from a menu or predictive fillin. All three components work well together and are preferred by users.

Second, this paper discusses principles for applying adaptive menus and predictive fillin to new application areas. Fields with a few, frequently repeated values are candidates for adaptive menus; functional dependencies indicate candidates for predictive fillin. Whether these characteristics can be learned at run-time is a topic for future research.

## Acknowledgments

Kerry Hersh Raghavendra provided the names used in Section 4. Apple Computer developed and supports Newton and the Newton ToolKit programming environment. The Newton AI group at WSU provided many useful comments on an earlier draft of this paper. Geoff Allen, Karl Hakimian, Mike Kibler, and the EECS staff provided a consistent and reliable computing environment. Anonymous reviewers of an earlier draft of this paper provided many (many) valuable suggestions. This work was supported in part by NASA under grant number NCC 2-794.